\documentclass{article}
\usepackage{PRIMEarxiv}
\pdfoutput=1
\usepackage[utf8]{inputenc} 
\usepackage[T1]{fontenc}    
\usepackage{hyperref}       
\usepackage{url}            
\usepackage{booktabs}       
\usepackage{amsfonts}       
\usepackage{nicefrac}       
\usepackage{microtype}      
\usepackage{lipsum}
\usepackage{fancyhdr}       
\usepackage{graphicx}       
\graphicspath{{media/}}     
\usepackage{amsthm,amsmath,amssymb}
\usepackage{mathrsfs}
\usepackage{subfigure}

\title{An Improved Strategy for Blood Glucose Control Using Multi-Step Deep Reinforcement Learning
\thanks{\textbf{*Weiwei Gu is the corresponding author.}} 
}

\author{
  Senquan Wang \\
  College of Information Science, \\
  Beijing University of Chemical Technology \\
  Beijing\\
  \texttt{senquanwang@buct.edu.cn}
   \And
  Weiwei Gu* \\
  College of Information Science, \\
  Beijing University of Chemical Technology \\
  Beijing\\
  \texttt{weiweigu@mail.buct.edu.cn} \\
}

\begin{document}
\maketitle

\begin{abstract}
Blood Glucose (BG) control involves keeping an individual's BG within a healthy range through extracorporeal insulin injections is an important task for people with type 1 diabetes. However, traditional patient self-management is cumbersome and risky. Recent research has been devoted to exploring individualized and automated BG control approaches, among which Deep Reinforcement Learning (DRL) shows potential as an emerging approach. In this paper, we use an exponential decay model of drug concentration to convert the formalization of the BG control problem, which takes into account the delay and prolongedness of drug effects, from a PAE-POMDP (\emph{Prolonged Action Effect-Partially Observable Markov Decision Process}) to a MDP, and we propose a novel multi-step DRL-based algorithm to solve the problem. The Prioritized Experience Replay (PER) sampling method is also used. Compared to single-step bootstrapped updates, multi-step learning is more efficient and reduces the influence of biasing targets. Our proposed method converges faster and achieves higher cumulative rewards compared to the benchmark in the same training environment, and improves the time-in-range (TIR), the percentage of time the patient's BG is within the target range, in the evaluation phase. Our work validates the effectiveness of multi-step DRL in BG control, which may help explore the optimal glycemic control measure and improve the survival of diabetic patients.
\end{abstract}


\section{Introduction}

Diabetes profoundly affects human life and health, regardless of country, age, or gender, and is one of the leading causes of death and disability worldwide \cite{ong2023global}. From 1990 to 2021, the age-standardized prevalence of diabetes increased by 90.5 \% globally, with increases of more than 100 \% in several regions, and it is projected that by 2050, there will be 1.31 billion people with diabetes worldwide \cite{ong2023global}. Furthermore, people with diabetes have more than twice the normal risk of early death, resulting in an estimated 150-500 million deaths around the world each year, while generating approximately 12\% of health expenditure (\$966 billion) \cite{tejedor2020reinforcement, federation2021idf}. The rising prevalence and serious health and economic hazards have attracted the attention of scientists around the globe, and as a result, the number of studies on diabetes is increasing. 

The pancreas of a diabetic does not produce or produces very little insulin, or the insulin produced is not used efficiently, leading to high BG and a variety of life-threatening complications such as cardiovascular disease, nerve damage, kidney damage, lower limb amputations, and eye disease leading to decreased vision and even blindness \cite{federation2021idf}. BG control is their basic treatment, as well as the basis for preventing and treating diabetic complications. Patients mainly maintain the stability of BG by injecting insulin. However, this traditional self-management is usually cumbersome and challenging, as it requires patients to measure their BG levels several times a day, while they suffer from many of the aforementioned complications \cite{tejedor2020reinforcement}.


To reduce this burden on diabetic patients and better help them with BG control, the Artificial Pancreas (AP) system has been proposed as an effective and safe new approach to BG management and has become the focus of diabetes research \cite{bekiari2018artificial}. The  AP consists mainly of three parts: a Continuous Glucose Monitoring (CGM) system, an insulin pump, and BG control algorithm. The closed-loop control algorithm is the core of the AP and continues to evolve. Previously, three main types of algorithms, namely the Proportional Integral Derivative (PID) control algorithm, Model Predictive Control (MPC) algorithm, and Fuzzy Logical (FL) control algorithm have been used in closed-loop BG control research. The PID controller is reactive and is unable to react fast enough to meals \cite{fox2020deep,garg2017glucose}. MPC builds systems based on assumptions that may ignore some unknown variables resulting in reduced accuracy \cite{bothe2013use}. The FL approach to modeling expert regulation of insulin delivery based on approximation rules may not provide customized BG control strategies for patients \cite{boughton2021new}. Overall, these traditional reactive controllers based on instantaneous glucose changes have not completely eliminated the effects of the delay inherent in the AP system predisposing to postprandial hyperglycemia, and the variability in BG concentrations due to dietary intake, exercise, sleep, and stress has not yet been effectively modeled, making it difficult to achieve patient-individualized precise glycemic control \cite{tejedor2023evaluating}.


In recent years, the Reinforcement Learning (RL) algorithm has been recognized as one of the most promising approaches capable of replacing the traditional paradigm of controlling insulin infusion, and it can be used to design a fully closed-loop controller that provides a truly personalized BG control scheme based entirely on the patient's own data \cite{tejedor2020reinforcement}. The BG control problem, which focuses on deciding the insulin dose to be injected into the patient at each moment, is a time-series decision-making problem, and RL algorithms are classical approaches for solving such sequence decision-making problems. RL makes decisions based on observed states, requiring neither a detailed description of the environment nor a large amount of labeled data for training as supervised learning does, and can learn directly on real data or interact with dynamic systems represented by mathematical models to automatically derive personalized dosing regimens for patients with life-threatening diseases \cite{bothe2013use,  ribba2020model}. The review by Tejedor et al. evaluates the use of RL methods to design algorithms for glycemic control in diabetic patients from 1990 to 2019 and predicts that RL algorithms will be used more frequently for BG control in the coming years \cite{tejedor2020reinforcement}. Fox et al. implemented automated BG control using three deep Q-networks that differ in architecture and state representation \cite{fox2019reinforcement}. Later, Fox et al. constructed the closed-loop BG control problem as a Partially Observable Markov Decision Process (POMDP), then used the Soft Actor-Critic (SAC) algorithm to learn a strategy for BG control, taking into account the noise of the CGM sensors and unobserved meal assumptions, and emphasized the flexibility of the RL methodology by experimentally comparing it with non-reinforcement learning algorithms \cite{fox2020deep}. Nordhaug Myhre et al. used a policy gradient RL approach for automated BG control, achieving performance that outperforms even state-of-the-art MPC \cite{nordhaug2020silico}. Zhu et al. also used the SAC algorithm and introduced food information through the Gate Recurrent Unit (GRU) \cite{zhu2021reinforcement}. Di Felice et al. compared the performance of two algorithms, DDPG and SAC, on a BG control task, with the latter achieving superior performance \cite{di2022deep}. Basu et al. considered the delay and prolongedness of drug effects and made improvements based on the DQN algorithm, finally proposing Effective-DQN algorithm to achieve individualized control of BG \cite{basu2023challenges}.


The combination of RL and deep learning with outstanding representational capabilities is being brought to its greater potential to scale to previously intractable problems \cite{arulkumaran2017brief}. One of the most exemplary representatives of DRL is the Deep Q-Network (DQN)\cite{mnih2015human}, which can learn successful policies directly from high-dimensional inputs using end-to-end RL and is the first artificial agent capable of learning to excel in a variety of challenging tasks. To date, many extensions have been proposed to improve the learning speed or stability of DQN. For example, Double DQN \cite{van2016deep} (DDQN), PER \cite{schaul2015prioritized}, Dueling network architecture \cite{wang2016dueling}, multi-step Q-learning \cite{sutton1988learning, sutton1999reinforcement}, Distributional Q-learning \cite{bellemare2017distributional}, and Noisy DQN \cite{fortunato2017noisy}, etc., all of the above components are integrated into a single integration agent is called Rainbow \cite{hessel2018rainbow}. The use of many DQN extensions can be seen in drug dosing problems, and they can often obtain more promising results than the classical DQN algorithm. For example, Raghu et al. combined the DDQN, Dueling DQN framework, and PER methods to produce the Dueling Double-Deep Q-Network (D3QN) to learn optimal sepsis treatment strategies \cite{raghu2017continuous}. Lopez-Martinez et al. also used D3QN to learn optimal opioid administration strategies for the management of critical care pain \cite{lopez2019deep}. Likewise, the use of DQN extensions can be found in studies of glucose control problems, e.g., Zhu et al. conducted a study of monohormonal (insulin) and dual-hormonal (insulin and glucagon) administration for glucose control using DDQN as a dosing strategy \cite{zhu2020basal}. Emerson et al. explored three approaches, Batch Constrained Deep Q-learning (BCQ), Conservative Q-learning (CQL) and Twin Delayed DDPG with Behavioral Cloning (TD3-BC), for the management of BG, demonstrating the potential of offline RL to create safe and sample-efficient glycemic control strategies in patients with type 1 diabetes. Several deep q-learning algorithms were thoroughly tested and compared by Tejedor et al \cite{tejedor2023evaluating}. However, they did not consider the prolongedness of the action effects and the multi-step learning approach.

Most previous articles using RL for BG control have used single-step reward bootstrapping; however, the target for single-step updating may be biased and lead to inefficient learning. Multi-step learning can reduce this bias and learn efficiently, yielding better performance. Therefore, in this article, we use a DQN algorithm extended with a multi-step learning approach to implement personalized BG control for virtual type 1 diabetic patients and incorporate the PER sampling method to improve learning efficiency. We also consider the delayed and prolonged drug effects and use a drug concentration decay model to convert the BG control problem from a PAE-POMDP to a MDP. In the modified Simglucose environment, our proposed improved strategy outperforms the benchmark in both the training and evaluation phases. The remainder of this paper is organized as follows: section 2 presents background knowledge of the decision process and the algorithmic mechanics of DQN and its several variants. Section 3 provides the technical approach. Section 4 shows and analyzes the results of training and evaluation. Finally, conclusions and promising future directions are offered in section 5.

\section{Background}

\subsection{Decision Process}
RL aims to learn the optimal strategy for executing the action from the delayed reward obtained through interaction with the environment and trial-and-error exploration. In general, at time \emph{t}, the agent obtains an observation \emph{$s_t$} from the environment and then takes an action \emph{$a_t$} according to policy $\pi$. Subsequently, the environment is affected by the agent's action and changes accordingly into the next state \emph{$s_{t+1}$}, and according to the reward function, the agent receives the corresponding reward \emph{$R_{t+1}$}. The MDP, a typical formal framework in RL, can describe sequential decision-making processes like the above. It is represented by a 5-tuple $(\mathcal{S}, \mathcal{A}, \mathcal{R}, \mathcal{P}, \gamma)$, where $\mathcal{S}$ is the set of states, $\mathcal{A}$ is the set of actions taken by the agent, $\mathcal{P}$ is the state transfer function that defines the dynamics of the environment, $\mathcal{R}$ is the reward function, and $\gamma$ is the discount factor. The MDP {process?} {has} obeys the Markov property that future states and rewards depend only on current states and actions and are independent of history. The ultimate goal of RL is finding a strategy for maximizing the cumulative reward from interacting with the environment over a series of discrete time steps\cite{sutton1999reinforcement}.

However, in reality, it is difficult to obtain complete state measurements, like during drug dosing, where the prolonged effects of historical drugs are not easily measured, and future drug effects depend on previous drug doses and their effects, which makes the problem no longer consistent with the Markov property and generalizes to POMDP. In POMDP, agents can only observe partial state measurements and need to infer the state through observation before acting \cite{basu2023challenges}. Thus, compared to the representation of {the} MDP, the POMDP has an additional set of observations $\Omega$ and an observation function $\mathcal{O}$ that denotes the probability of obtaining an observation \emph{o} based on a new state \emph{$s'$} and action \emph{a}, which is ultimately represented as a 7-tuple $(\mathcal{S}, \mathcal{A}, \mathcal{R}, \mathcal{P}, \gamma, \Omega, \mathcal{O})$. In this case, the goal of the agent remains to obtain the maximum cumulative reward. POMDPs are common in various domains such as games, robotics, natural language processing, transportation, communication and networking, and industry, which can be converted into belief MDPs for processing \cite{xiang2021recent} or by combining RL with recurrent neural networks, thus effectively introducing historical information for processing. \cite{meng2021memory}.

In this paper, we are concerned with PAE-POMDPs, in which the effect of an action not only affects the state at the next time step but also has a different impact on the subsequent time steps. From a forward perspective, it is assumed that the action \emph{$a_t$} taken at the current moment \emph{t} affects the future state for $\kappa$ time steps, where $\kappa$ depends on the environment and the amplitude of the action. From a backward perspective, the state \emph{$s_t$} at the current moment \emph{t} is affected by the superposition of the previous $\kappa$ time steps. As a result, the state transfer function \emph{P($s_{t+1}$}|$s_t$,$a_t$) of traditional MDPs is no longer valid and is extended to \emph{P($s_{t+1}$}|$s_t$,$(a_{t-k})_{k=0}^{\kappa}$). While the reward function of PAE-POMDPs remains as \emph{R($s_t$,$a_t$,$s_{t+1}$}). A more detailed description can be found in \cite{basu2023challenges}.

\subsection{DQN and its extensions}
Deep Q-Network (DQN), a model-free RL algorithm, belongs to the class of Q-learning algorithms. \emph{Q($s_t$,$a_t$)} describes the quality of the action \emph{$a_t$} made by the agent in state \emph{$s_t$}, referred to as the value function, which is learned through temporal-difference updates.
\begin{equation}
Q(s_t,a_t) = Q(s_t,a_t) + \alpha[r_t + \gamma \max\limits_{a} Q(s_{t+1},a) - Q(s_t,a_t)],
\end{equation}
where $\alpha$ is the learning rate. In DQN, a neural network is used to fit the value function \emph{Q($s_t$,$a_t$)}. The network parameters are updated by gradient descent to minimize the mean square Bellman error. 
\begin{equation}
L_i(\theta_i)=\mathbb{E}_{s_t,a_t~\pi}[(r_t+\gamma\max\limits_{a_{t+1}}Q(s_{t+1},a_{t+1};\theta_{i-1})-Q(s_t,a_t;\theta_i))^2],
\end{equation}
where $\pi$ is the policy and $\theta_i$ are the parameters of the Q-network at iteration \emph{i}. Eventually, it will converge to obtain the optimal value function \emph{$Q^*$}, which can extract the optimal policy \emph{$\pi^*(s_t$)=}$arg\max_{a\in\mathcal{A}}Q(s_t,a)$.

The DQN has many extensions that can improve its speed or stability. 

\paragraph{DDQN} uses two neural networks to fit \emph{Q($s_t$,$a_t$)}, where the target Q-value is not determined directly by the target network but with the help of the main network. This reduces the overestimation of the Q-value and yields better performance \cite{van2016deep}.

\paragraph{Dueling DQN} adds stability to the optimization by transforming the two neural networks, both previously used to fit \emph{Q($s_t$,$a_t$)}, into separate networks for evaluating the merits of states and actions. This architecture leads to better strategy evaluation in the presence of many similar-valued actions \cite{wang2016dueling}.

\paragraph{PER} argues that the amount an agent can learn from a sample determines the importance of that sample. Where the amount of learning can be measured by the magnitude of the Bellman error $\delta$, samples with large $\delta$ are more important and require more learning. Therefore important samples are sampled frequently to increase training efficiency \cite{schaul2015prioritized}.

\paragraph{Multi-step learning} using n-step cumulative discount rewards for differential updating enables faster learning \cite{hessel2018rainbow}. Furthermore, goals bootstrapped through a single step can be biased, and using longer trajectories that incorporate more observations of future rewards is one way to reduce such biases \cite{hernandez2019understanding}.

\paragraph{NoisyNet-DQN} replaces the linear layer in the DQN with a noise layer whose weights and biases are perturbed by a parametric noise function \cite{fortunato2017noisy}. It combines deterministic and noisy streams, thus allowing agents to perform state-conditioned exploration in a self-annealing form \cite{hessel2018rainbow}.

\paragraph{Distributed DQN} learns an approximate full distribution of Q values, which mitigates the effects of learning from a nonstationary strategy, and uses a distributed Bellman operator that preserves multimodality in the distribution of values, so more robust learning can be achieved compared to predicting individual Q values \cite{bellemare2017distributional}.

\paragraph{Rainbow DQN} integrates all of the aforementioned extensions, which improve data efficiency and overall performance. In terms of median human-normalized performance metrics across 57 Atari games, it outperforms any baseline by a wide margin. \cite{hessel2018rainbow}.

\section{Methods}

\begin{figure}
  \centering
  \includegraphics[width=1\linewidth]{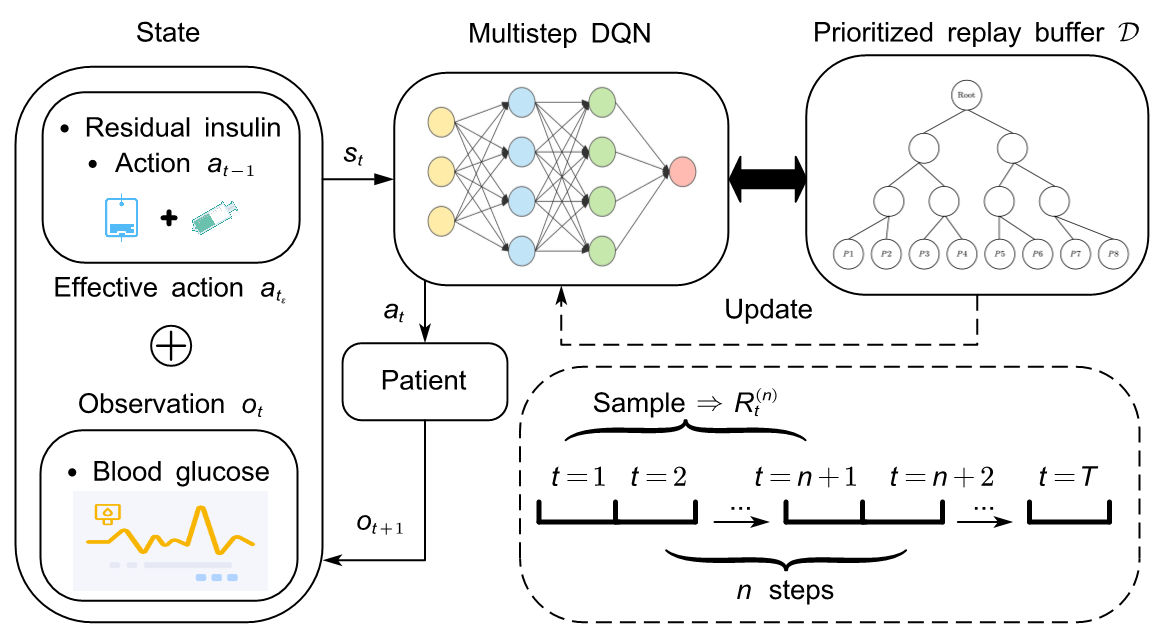}
  \caption{Schematic diagram of the algorithm for the BG control problem.}
  \label{fig1}
\end{figure}

\subsection{Converting PAE-POMDP into MDP}
In order to use DRL algorithms such as DQN directly to solve the drug dosing problem, the PAE-POMDP needs to be converted to a traditional MDP. Using an exponential decay model of drug concentration over time allows for easy conversion of PAE-POMDP to MDP, thus avoiding the complexity of recording and encoding historical actions and reducing both computational time and model size \cite{basu2023challenges}. The model assumes that historical action effects decay at a fixed rate $\lambda$ and that action effects are independently capable of being cumulative. Formally, we call this cumulative action effect as \emph{effective action}, $a_{t_{\varepsilon}}$, and define it as:
\begin{equation}
    a_{t_{\varepsilon}} = 
        \begin{cases}
            0,  & \text{if $t=0$} \\
            \lambda a_{(t-1)_{\varepsilon}} + a_{t-1}, & \text{otherwise}
        \end{cases}
\end{equation}
where $\lambda \in (0,1)$ is an assumed exponential decay rate that needs to be adjusted according to the drug and patient metabolism. Note that setting $\lambda=0$ returns to the classic RL scenario, whereas setting $\lambda=1$ implies that the action effects of the history have a permanent effect on subsequent time steps. During the iterative interaction of the agent with the environment, the computed \emph{effective actions} are incorporated into the current state as known information, thus allowing the agent to take into account the dynamic impact of the historical actions on subsequent time steps, i.e., the observation of the agent at the current time step $t$ can be defined in terms of the current state and all the actions in the past:$o_{t+1} \sim \mathcal{O}(\bullet|s_t,a_{\textless t})$. By augmenting the original incomplete observation, the POMDP is restored as an MDP $\mathcal{M}_{\varepsilon}(\mathcal{S}_{\varepsilon}, \mathcal{A}, \mathcal{R}, \mathcal{P}, \gamma)$, where the states in $\mathcal{S}_{\varepsilon}$ are defined as $s_{t_{\varepsilon}}=(s_t,a_{t_{\varepsilon}})$ and the restored MDP is only altered in the state compared to the conventional MDP.

\subsection{Prioritized Experience Replay (PER)}
In traditional experience replay, state transition tuples are placed into the replay buffer as training samples without any distinction, and each training sample is sampled with the same probability, which prevents some important samples from getting the attention they deserve. Moreover, when the replay buffer capacity is exceeded, the samples put in earlier will be deleted, and those samples worth learning may be deleted. Most of the storage space is occupied by unimportant training samples, which leads to an imbalanced training problem and severely reduces the learning efficiency \cite{yuan2021prioritized}.

In this work, we use PER for efficient sampling. First, we use the structure of a summing tree as a replay buffer to collect samples because the tree structure is more computationally efficient than the queue structure when sampling values in order of magnitude. Then, an importance priority $p_t$ is assigned to each sample, which is related to the absolute TD error of the final computation of the sample:
\begin{equation}
p_{t} \propto\left| R_{t+1}+\gamma_{t+1} \operatorname* {m a x}_{a^{\prime}} q_{\overline{{{\theta}}}} ( S_{t+1}, a^{\prime} )-q_{\theta} ( S_{t}, A_{t} ) \right|^{\omega},
\end{equation}
where $\omega$ is a hyper-parameter that determines the shape of the distribution. Specifically, new transitions are inserted into the replay buffer with maximum priority.

\subsection{Multi-step Q-Learning for BG control problems}
The traditional DQN approach introduced in Section 3.2 uses a single-step reward for updating, which is easy to implement and can be directly applied to solving high-dimensional state-space MDPs. However, for BG control problems, due to the delayed and prolonged effects of insulin action, the impact of taking the action of injecting insulin may not be immediately altered and reflected in the status and reward of the next time step, but rather affect rewards at multiple time steps later. Therefore, many updates are required to propagate rewards to relevant preceding states and actions, making sampling inefficient and leading to slower learning \cite{mnih2016asynchronous, qin2021solving}.

We address the above problem by adopting the multi-step learning approach, specifically using forward-view multi-step targets \cite{sutton1988learning}, that is, the truncated n-step return $R_t^{(n)}$ of the given state $S_t$:
\begin{equation}
    R_t^{(n)} = R_{t+1}+\gamma R_{t+2}+\ldots+{\gamma}^{n-1}R_{t+n} = \sum_{k=0}^{n-1}\gamma R_{t+k+1}\label{Rn}
\end{equation}
The action-value function \emph{Q($s_t$,$a_t$|$\theta$)} can then be updated by minimizing the alternative loss:
\begin{equation}
    L(\theta) = (R_t^{(n)} + {\gamma}^n\max\limits_{a'}Q(s_{t+n},a'|{\theta}') - Q(s_t,a_t|\theta))^2
\end{equation}
In particular, the buffer $\mathcal{D}$ sequentially stores tuples of state transitions $(s,a,r,s')$ generated by the interaction of the agent with the environment, and the rewards $r_{t+k} (k=1,\ldots,n)$ and state $s_{t+n}$ are obtained by taking samples of consecutive n-steps from $\mathcal{D}$ thereby obtaining $R_t^{(n)}$ according to equation (\ref{Rn}). The overall schematic of the solution to the BG control problem using the PER and multi-step learning approach is shown in Figure \ref{fig1}.

By applying multi-step learning, both the long-term and short-term effects of actions can be learned by regressing toward the exact reward rather than being bootstrapping from the target network $Q(s,a|{\theta}')$ \cite{qin2021solving}, which will significantly improve the learning efficiency for the BG control problem.

\section{Experiments and Results}

In this section, we validate the performance of the proposed multi-step learning approach that uses PER in solving the BG control problem. We first give the setup of the experimental environment, then present the setup of the algorithm, and finally show and analyze the training and evaluation results.

\subsection{BG control Environment}
We used the modified Simglucose simulator of Basu et al. \cite{basu2023challenges} in our experiment. The Simglucose \cite{Jinyu2018Simglucose} is the OpenAI Gym \cite{brockman2016openai} implementation of the UVA/Padova Type1 Diabetes Mellitus Simulator (T1DMS) \cite{dalla2009physical, man2014uva} approved by FDA. The simulator has 30 in-silico patients, including 10 children, 10 adolescents, and 10 adults. We randomly selected patient adult\#009 as the environment to train the RL agents as we wanted to get individualized BG control strategies.

In this case, the agent observes the continuous BG value of the patient in the simulated environment, which is used as the state in conjunction with the effective insulin injection action $a_{t_{\varepsilon}}$. The agent can take the action of injecting 0-5 discrete doses of insulin with the aim of controlling the simulated patient's BG to stabilize it within a normal range for as long as possible. The environment is set where the starting BG value of the virtual patient in each episode is uniformly randomly selected from the range (150,195), and the episode is terminated when the BG is below 70 (hypoglycemic death) or above 200 (hyperglycemic death). The segmentation of the virtual patient's BG condition is described in Table \ref{table1}. To this end, we used zone rewards \cite{basu2023challenges} to encourage the agent to maximize the time the patient's BG is in the target zone with as few insulin doses as possible and penalized the agent for hyperglycemia, hypoglycemia and performance with excessive insulin doses. Formally, zone rewards are expressed as:

\begin{equation}
r_{t} ( s_{t-1}, a_{t}, s_{t} )=r_{t, s t a t e} ( s_{t-1}, s_{t} )-r_{t, a c t i o n} ( a_{t} ) ,
\end{equation}

where $r_{t, s t a t e} ( s_{t-1}, s_{t} )$ called state reward incentivizes BG to be within a healthy target range as follows:
\begin{equation}
    r_{t, s t a t e} ( s_{t-1}, s_{t} )=
        \begin{cases}
            -100,  & \text{$s_t<70$ or $s_t>200$ (episode termination)} \\
            -1,  & \text{$s_t<100$ and $s_t-s_{t-1}<0.5$ (hypoglycemia)} \\
            -1,  & \text{$s_t>150$ and $s_t-s_{t-1}>0.5$ (hyperglycemia)} \\
            10,  & \text{$100 \leq s_t \leq 150$ (target BG)}
        \end{cases},
\end{equation}

and $r_{t, a c t i o n} ( a_{t} )$ is a penalty term penalizing the agent for overdosing:
\begin{equation}
r_{t, a c t i o n} ( a_{t} )=0. 1 * a_{t}^{2}.
\end{equation}

\begin{table}[h]
 \caption{BG segmentation}
  \centering
  \begin{tabular}{ccccc}
    \toprule
    hypoglycemic death & hypoglycemic  & Euglycemia & hyperglycemic & hyperglycemic death \\
    \midrule
    $\leq70$ & (70,100) & [100, 150] & (150,200] & $>200$ \\
    \bottomrule
  \end{tabular}
  \label{table1}
\end{table}

\subsection{Algorithm Setup}
We use Effective-DQN \cite{basu2023challenges}, which performs well in the BG control task considering the delay and prolongedness of actions, as a benchmark and then compare the proposed improved method with Effective-DQN. In addition, we conducted ablation experiments to analyze the effects of prioritized experience buffer and multi-step learning on overall performance, respectively.

The hyperparameters of the proposed improved algorithm are provided in Table \ref{table2}. The hyperparameters of the benchmark Effective-DQN are the same as those of the improved algorithm, except for the multi-step n and PER-related parameters.

\begin{table}[h]
 \caption{Hyperparameters of multi-step learning with PER}
  \centering
  \begin{tabular}{cccc}
    \toprule
    Hyperparameter & value  & Hyperparameter & value \\
    \midrule
    Number of episode & 10000 & Number of explore $n$ & 1000 \\
    Replay buffer size & $10^6$ & Batch size & 512 \\
    Discount factor $\gamma$ & 0.999 & Learning rate $\eta$ & 0.001 \\
    Greedy decay & 500 & Action decay rate $\lambda$ & 0.95 \\
    Greedy maximum $\overline{\epsilon}$ & 0.9 & Greedy minimum $\underline{\epsilon}$ & 0.05 \\
    Priority exponent $\omega$ & 0.5 & Priority weight $\beta$ & 0.4 \\
    Multi-step $n$ & 16 & Hidden size & 256 \\ 
    \bottomrule
  \end{tabular}
  \label{table2}
\end{table}

\subsection{Results}

\begin{figure}
  \centering
  \includegraphics[width=1\linewidth]{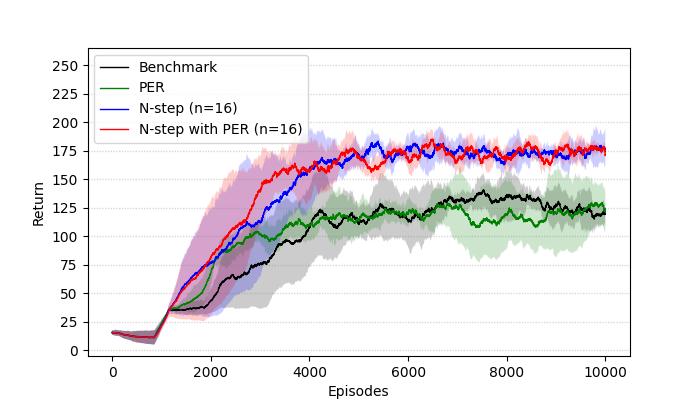}
  \caption{Cumulative discounted sum of rewards during the training process.}
  \label{fig2}
\end{figure}

\paragraph{Training.}
Figure \ref{fig2} shows the cumulative discounted sum of rewards after each training episode for our proposed method and benchmark, along with the use of PER alone and multi-step learning alone. In this case, the solid line is the mean after multiple runs and smoothing, and the shadows indicate the standard deviation. Here, we performed 5 times training for each algorithm by setting random seeds of 1-5.

From Fig.\ref{fig2}, we can see that our proposed approach, multi-step learning combined with the PER, achieves higher cumulative discount rewards than the benchmark in the simulated BG control environment, and the algorithm is able to converge faster. The model using PER alone performs comparably to the benchmark model but achieves higher cumulative rewards faster. Multi-step learning combined with PER is more stable than normal multi-step learning because it focuses on collecting those more important samples for training at a later stage of the training process, corresponding to fewer shadow areas on the graph.

\begin{figure}[htbp]
\centering
\subfigure[Benchmark]{\label{fig:subfig:a}
\includegraphics[width=0.48\textwidth]{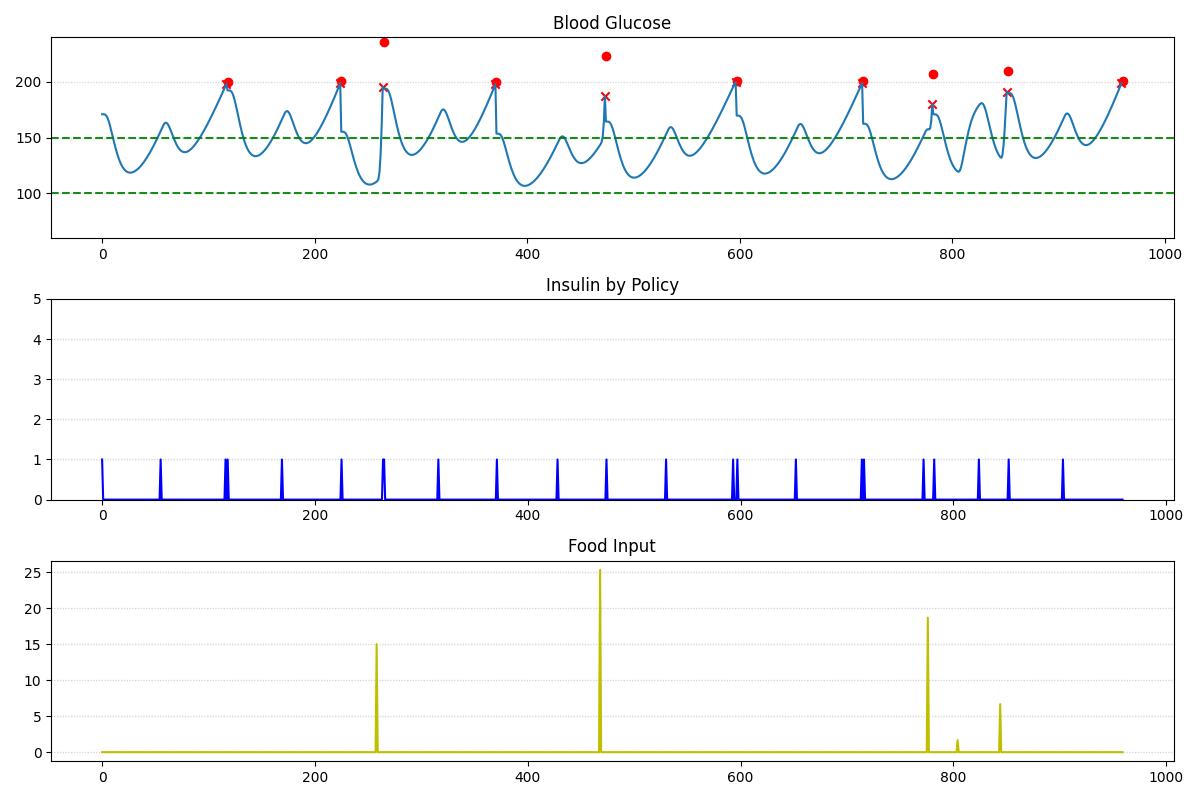}}
\subfigure[Our approach]{\label{fig:subfig:b}
\includegraphics[width=0.48\textwidth]{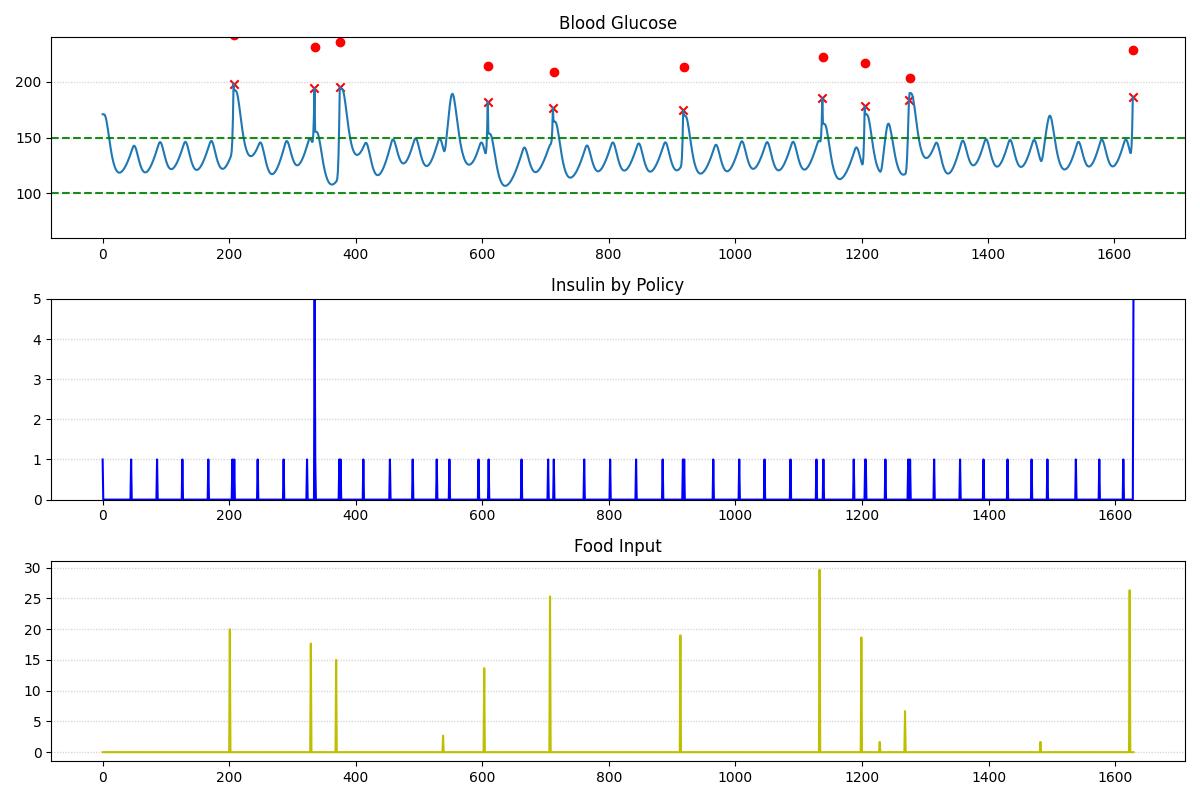}}
\caption{Policy visualization.}
\label{fig3}
\end{figure}

\paragraph{Evaluation.}
After training, we tested the performance of the algorithms for each algorithm by randomly selecting the model trained with seed 3 from the 5 models trained. The same virtual patient environment was used for testing and training, with the difference that the 10 evaluations were set with seeds of 0, 10...100, which ensured that the starting BG values were the same for the different algorithms tested. The test results for each algorithm are shown in Figure \ref{fig3}.

Here, the scheduling of food intake is uniform for each agent in each episode, due to differences in the termination time of the episodes, making food intake timing look different. It can be observed that both strategies learn to perform the conservative action of administering the lowest dose of insulin due to the fact that the model takes into account the information of residual insulin in the bloodstream from previous administrations in order to avoid hypoglycemia triggered by too much administration and that there is a penalty for too much administration in the zone reward. In particular, the proposed improved multi-step learning strategy with PER is more sensitive to the increase of BG. It can make insulin administration more quickly than the baseline strategy, and frequent administration of small doses helps BG stay in the safe target region for more time. For transient increases in BG triggered by large doses of food intake, both agents chose to perform insulin administration around the food intake, with the surprise being that our model occasionally made an action to increase the dose of insulin administered, but the magnitude of the administered dose remained small. The reasons for this may be twofold: (1) the action space is small and even performing the maximum dose of administration does not avoid hyperglycemia triggered by large food intake. (2) it is difficult to train appropriate coping strategies because of the small number of samples of large food intake during the training phase and because the episodes were terminated by triggering hyperglycemia whenever there was a large amount of food intake.

\begin{figure}[htbp]
\centering
\subfigure[Benchmark]{\label{fig:subfig:a}
\includegraphics[width=0.48\textwidth]{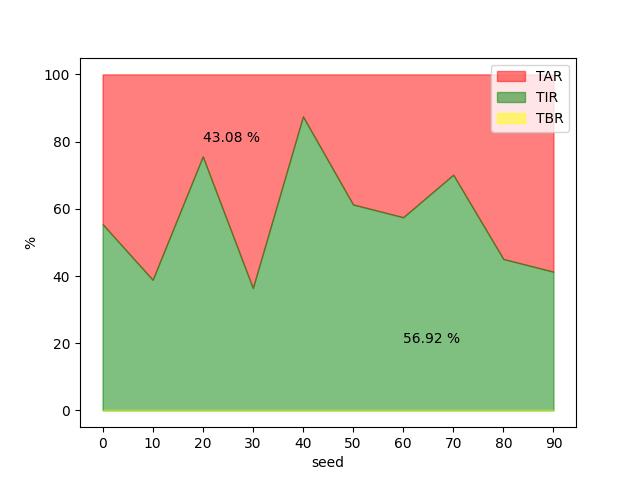}}
\subfigure[Our approach]{\label{fig:subfig:b}
\includegraphics[width=0.48\textwidth]{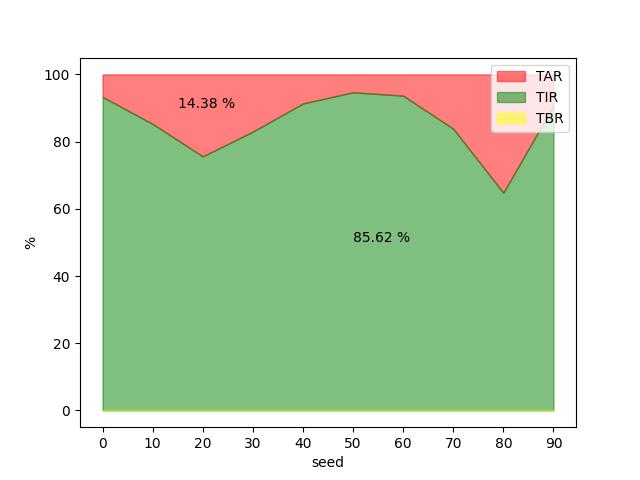}}
\caption{Time distribution of BG in different ranges.}
\label{fig4}
\end{figure}

Finally, we compare the test results of each model more visually through Figure \ref{fig4}. We show the TIR, the time-above-range (TAR), and the time-below-range (TBR) of the patient's BG under the control of the different strategies. The proposed improved strategy of multi-step learning has a target range time of 85.62\% of the total test time, which is 28.7\% higher than the benchmark strategy. 

\section{Discusion}
\paragraph{Conclusion.}In this paper, we propose a multi-step DRL algorithm to solve the BG control problem, which takes into account the delayed and prolonged effects of drugs and converts a PAE-POMDP into a MDP by using an exponential decay model of drug concentration. We use a deep network parameterization to approximate the action-value function, adopt PER and consider multi-step cumulative rewards for the training of the agents.10 episodes of BG control tests show that our proposed algorithm significantly outperforms the baseline model, Effective-DQN, in terms of TIR metrics of BG in patients. 
\paragraph{Limitations and future work.}In the experiments, we learned the optimal dosing strategy based only on the changes in the patient's BG values, but from the results, we were able to observe that almost all episodes were terminated due to a sudden increase in BG triggered by food intake. Exploring strategies to address this problem by introducing information on food intake may lead to more interesting insights in the future. In addition, the patient's BG level is affected by many other factors, such as exercise, sleep, and the effects of other medications, which need to be studied in more depth and detail in the future.

\bibliographystyle{unsrt}  
\bibliography{myref}

\end{document}